\documentclass[runningheads,a4paper]{llncs}

\usepackage{amssymb}
\usepackage{graphicx}

\usepackage{url}
\urldef{\mailsa}\path| {mina.rezaei, haojin.yang, christoph.meinel}@hpi.de|  
\newcommand{\keywords}[1]{\par\addvspace\baselineskip
\noindent\keywordname\enspace\ignorespaces#1}

\begin{document}

\mainmatter  

\title{Deep Neural Network with l2-norm Unit for Brain Lesions Detection}

\titlerunning{Convolutional Neural Networs for Lesions Detection in Brain MRI}

%
%
\author{Mina Rezaei, Haojin Yang, Christoph Meinel}
\authorrunning{Convolutional Neural Networs for Lesions Detection in Brain MRI}

\institute{Hasso Plattner Institute,\\
Prof.-Dr.-Helmert-Straße 2-3, 14482 Potsdam, Germany\\
\mailsa\\ }

%
%
\toctitle{Lecture Notes in Computer Science}
\tocauthor{Authors' Instructions}
\maketitle

\begin{abstract}
Automated brain lesions detection is an important and very challenging clinical diagnostic task, because the lesions have different sizes, shapes, contrasts and locations.
Deep Learning recently shown promising progresses in many application fields, which motivates us to apply this technology for such important problem.
In this paper we propose a novel and end-to-end trainable approach for brain lesions classification and detection by using deep Convolutional Neural Network (CNN).
In order to investigate the applicability, we applied our approach on several brain diseases including high and low grade glioma tumor, ischemic stroke, Alzheimer diseases, by which the brain Magnetic Resonance Images (MRI) have been applied as input for the analysis.
We proposed a new operation unit which receives features from several projections of a subset units of the bottom layer and computes a normalized l2-norm for next layer.
We evaluated the proposed approach on two different CNN architectures and number of popular benchmark datasets.
The experimental results demonstrate the superior ability of the proposed approach.
\keywords{Multimodal CNN, l2-norm unit, Brain lesion detection and localization}
\end{abstract}

\section{Introduction}

Annually in the United State alone 24,000 adult and 4,830 children will be diagnosed as new cases of brain cancer.
A lot of people have died due to brain tumor, multiple sclerosis, ischemic stroke and Alzheimer diseases~\footnote{\url{http://www.cancer.net/cancer-types/brain-tumor/statistics}}. 
Medical imaging is an important tool for brain diseases diagnosis in case of surgical or chemical planning. Magnetic Resonance Imaging (MRI) can provide rich information for premedication and surgery medication, 
which is extremely helpful for evaluating the treatment and lesion progress. 
However the raw data extracted from MR images is hard to be directly applied for diagnosis due to the large amount of the data.
An accurate brain lesion detection and classification algorithm based on MR images might be able to improve the prediction accuracy and efficiency, 
that enables a better treatment planning and optimize the diagnostic progress.
As mentioned by Menze et al.~\cite{Menze2014}, the number of clinical study for automatic brain lesion detection has grown significantly in the last several decades. 
Some brain lesions such as ischemic strokes, or even tumors can appear with different shapes, inappropriate sizes and unpredictable locations within the brain. 
Furthermore, different types of MRI machines with specific acquisition protocols may provide MR images with a wide variety of gray scale representations on the same lesion cells. 
Recent research has shown strong ability of Convolutional Neural Network (CNN) for learning hierarchical representation of image data without requiring any effort to design handcrafted features~\cite{lecun2015deep,szegedy2015going,gulcehre2014learned}. 
This technology became very popular in computer vision society for image classification~\cite{krizhevsky2012imagenet,he2016deep}, object detection~\cite{szegedy2013deep,girshick2014rich,RenHG015}, medical image classification~\cite{doi2254195,el2017brain} and segmentation\cite{ronneberger2015u,dai2016instance}. 
As mentioned by LeCun et al. in \cite{lecun2015deep}: different layers of a network are capable of different levels of abstraction, and capture different amount of structures from the patterns present in the image. 

In this work we investigate the applicability of CNN for brain lesions detection. 
Our goal is to perform localization and classification of single as well as multiple anatomic regions in volumetric clinical images from various image modalities. 
To this end we propose a novel framework based on CNN with l2-norm unit.
A detailed evaluation on parameter variations and network architectures has been provided.
We show that l2-norm operation unit is robust to the error variations in the classification task 
and is able to improve the prediction result. 
We conducted experiments on a number of brain MRI datasets, which demonstrate the excellent generalization ability of our approach.
The contribution of this work can be summarized as following:

\begin{itemize}
\item We propose a robust solution for brain lesions classification.
We achieved promising results on four different brain diseases (The overall accuracy is over 95\%).
\item We applied multiple MRI modalities as network input, and this improved the dice coefficient up to 30\% on ISLES benchmark.
\item We implemented l2-norm unit in Caffe~\cite{DBLP:journals/corr/JiaSDKLGGD14} framework for both CPU and GPU computation.
The experimental results demonstrate the superior ability of l2-norm in various tasks.
\end{itemize}

The rest of the paper is organized as follows: Chapter~\ref{methodology} describes the proposed approach, Chapter~\ref{exprimental} presents the detailed experimental results.
Chapter~\ref{conclusion} concludes the paper and gives an outlook on future work.

\section{Methodology} \label{methodology}

In this chapter we will describe our deep network for classification and detection task in detail.
The core techniques applied in our approach are depicted as well.
 
In the recent deep learning context, a deep neural network can be built driven by two principles: Modularity and Residual learning.
Modularity is a set of repeatable smaller neural network unit which enables the learning of high-level visual representations.
The bottleneck module of the Inception architecture~\cite{DBLP:journals/corr/SzegedyVISW15} and the corresponding units in VGG-Net~\cite{simonyan2014very} can be considered as typical examples. 
In such networks the wide and depth have been significantly increased. 

\begin{figure} [!t]
\includegraphics[width=0.9\textwidth]{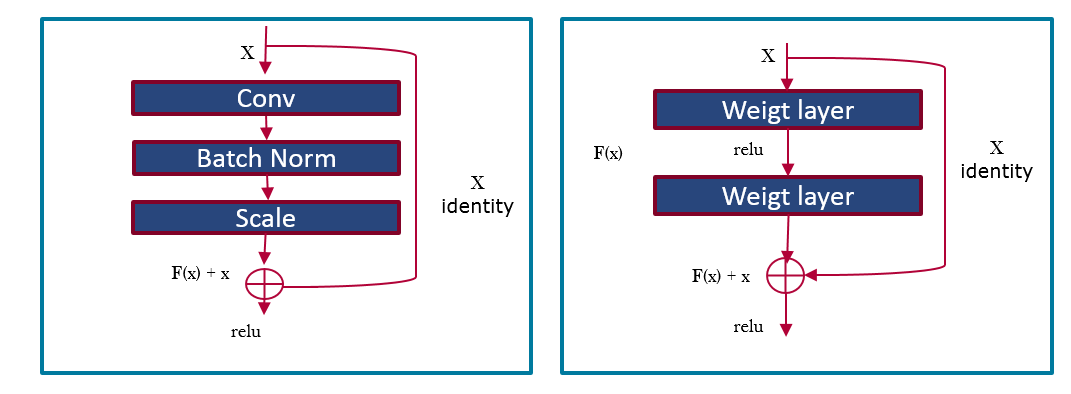}
\centering
\caption{Exemplary residual building block. 
The block on the left side shows a vanilla residual block, where the one on the right side is a dense block applied in our classification network.}
\label{fig_residual_units}
\end{figure}

On the other hand residual learning~\cite {he2016deep} considers new way to each layer.
Every consequent layer is responsible for, in effect, fine tuning the output from a previous layer by just adding a learned ``residual'' connection to the input. 
This essentially drives the new layer to learn something different from what the input has already encoded. 
Another important advantage is that such residual connections can help in handling gradient vanishing problem in very deep networks~\cite{he2016deep}.
Figure~\ref{fig_residual_units} shows an exemplary residual building block, where $F(x) + x$ denotes the element-wise addition of the original input and the residual connection.
The block on the left depicts vanilla residual unit proposed by He et al.~\cite{he2016deep}, where the one on the right side is a dense block that we utilize in our classification network.

\subsection{l2-norm Unit}

\begin{equation}
\label{eq_l2norm_unit}
\mid X_{i,j} \mid=
  \left[ {\begin{array}{cccc}
   x_{1,1} & x_{1,2} & ... & x_{1,j}  \\
   x_{2,1} & x_{2,2} & ... & x_{2,j}  \\
   ... & ... & ... & ...  \\
   x_{i,1} & x_{i,2} & ... & x_{i,j}  \\
  \end{array} } \right]
\end{equation}

\begin{equation}
\label{eq_l2norm_forward}
Forward: \mid x_{i,j} \mid =\sqrt[2] { \sum { {x_{i,j}}^{2}} }
\end{equation}

\begin{equation}
\label{eq_l2norm_backward}
Backward: {\partial \mid x_{i,j} \mid}
   = \frac{n \partial (\sum {x_{i,j}})  } {2 \sqrt[2] { \sum {x_{i,j}}^{2} } }
\end{equation}

In linear algebra, the size of a vector \textit{v} is called the norm of \textit{v}. 
The two-norm (also known as the l2-norm, mean-square norm, or least-squares norm) of a vector \textit{v} is defined by Equation~\ref{eq_l2norm_forward}. 
Assume we have a 2D matrix $X_{i,j}$ (cf. Equation~\ref{eq_l2norm_unit}) which is the output of the specific patch of $a_{i,j}$ from the first convolution layer.
Then for each item in feed forward or backward pass we calculate the l2-norm as described by Equation~\ref{eq_l2norm_forward} and \ref{eq_l2norm_backward}. 
We consider l2-norm operation as a pooling function and apply it to reduce the dimension of the learned representations, which is able to obtain better generalization ability. 
For example in the classification task an input volume of size $224\times224\times64$ is pooled by l2-norm operator with filter size 2 and stride 2 into an output volume of size $112\times112\times64$.

\subsection{Brain Abnormality Classification}

\begin{figure}[t!]
\includegraphics[width=0.9\textwidth]{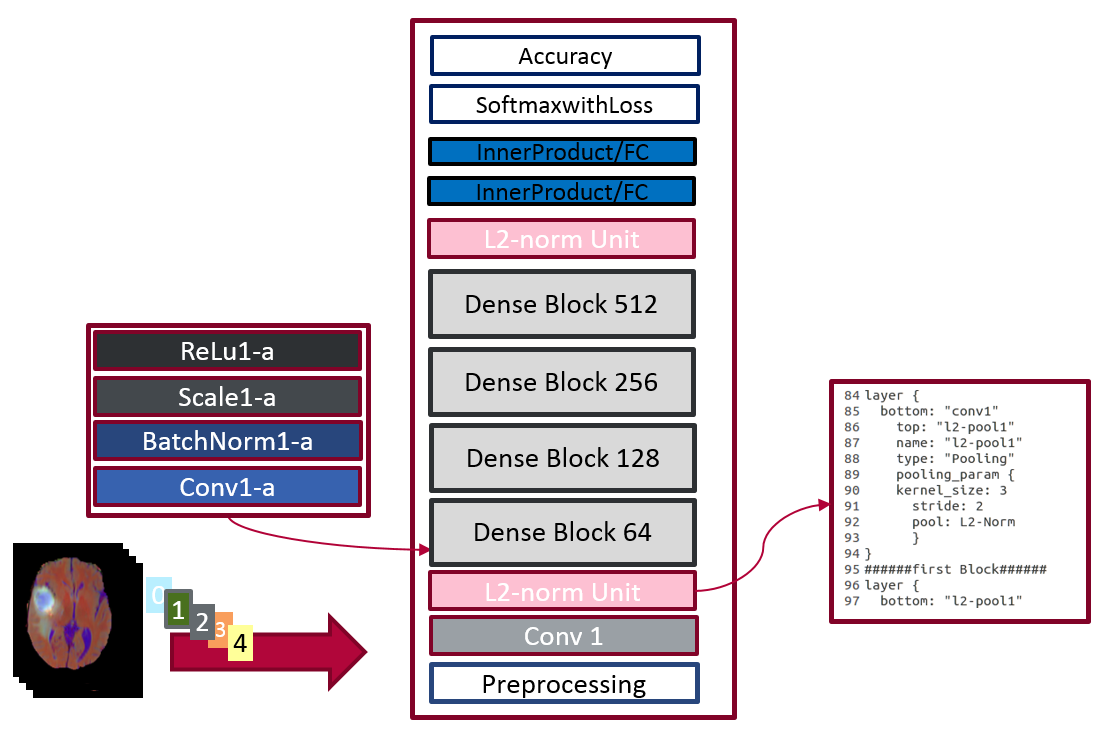}
\centering
\caption{Brain diseases classification architecture}
\label{fig_classification_arch}
\end{figure}  

Recently, ResNet (Deep Residual Network)~\cite{he2016deep} achieves the state-of-the-art performance in object detection and other vision related tasks. 
As mentioned above we explored the ResNet architecture with l2-norm unit for brain abnormality classification. 
Figure~\ref{fig_classification_arch} depicts the network architecture.
Our classification network takes 2D images with three channels, while each channel contains a gray scale copy with the same size and same plane from various MRI modalities with respective class label \textit{l=\{0,1,..,4\}}. 
Each gray scale copy extracted from T1, T1c and FLAIR of the same MRI categories has been mapped to the Red, Green and Blue channels of a standard image container, respectively. 
The proposed network strongly inspired by vanilla ResNet block depicted by Figure~\ref{fig_residual_units}.

As shown in Figure~\ref{fig_classification_arch}, we apply l2-norm operation after the first convolution layer and before the first inner product layer.
In the experiments we observed that the l2-norm layer performs a similar effect as a pooling operator, which reduces the spatial size of the feature representations and extracts features that are not covered by standard pooling operators.
This allows the network to learn more distinguished feature information such as variance from the data stream, which could improve the overall generalization ability of the model.

\subsection{Brain Lesions Detection}

\begin{figure}[t!]
\includegraphics[width=0.9\textwidth]{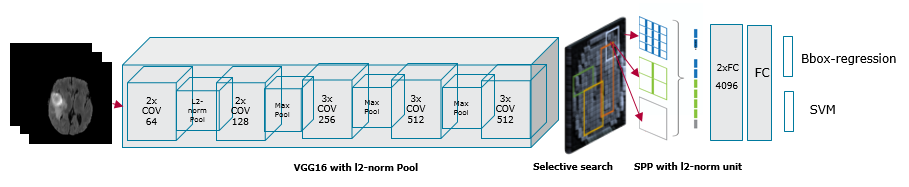}
\centering
\caption{proposed architecture with 16 convolutions and l2-norm unit for recognition and localization of  brain lesion}
\label{fig_detection_arch}
\end{figure}

Unlike image classification, object detection extracts location and region information of a target object within an image. 
Figure~\ref{fig_detection_arch} represents our network for brain abnormality detection.
In our workflow, we extract and apply multiple modalities from MRI images, where the images are sampled in 2D slices from the axial, coronal and sagittal view with various sizes.
Inspired by Fast R-CNN network~\cite{Girshick15}, we build our CNN network based on VGG-16~\cite{simonyan2014very} style architecture as the feature extractor. 
Instead of using max-pooling and spatial max-pooling we place the l2-norm unit after the second convolution(conv1-2) layer and before the first fully connected (inner product) layer respectively.
We utilize selective search~\cite{Uijlings13} to generate object proposals, which is a set of object bounding boxes.
The proposal sampling process is performed on top of dense feature layer after layer~\texttt{conv5-3}.
We confirm the suggested solution by Girshick et al.~\cite{Girshick15} to come over on heterogeneous collection of computed proposals and divide them into a pyramid grid of sub-windows.
Here three pyramid levels \textit{$4\times4, 2\times2, 1\times1$} and l2-norm ``pooling'' have been applied in each sub-window to generate the corresponding output grid cell.
Subsequently each output feature vector is further fed into a sequence of fully connected layers, which is followed by two sibling output layers: the SVM (Support Vector Machine) classifier for object class estimation~\cite{liu2008multi}, and the bounding box regression layer to calculate the loss of proposed object bounding boxes.
The overall training is performed in the supervised manner, and the loss of the whole network sums losses from both object classification and bounding box regression.
 
\section{Exprimental Results}  \label{exprimental}
\begin{figure} [!t]
\includegraphics[width=0.6\textwidth]{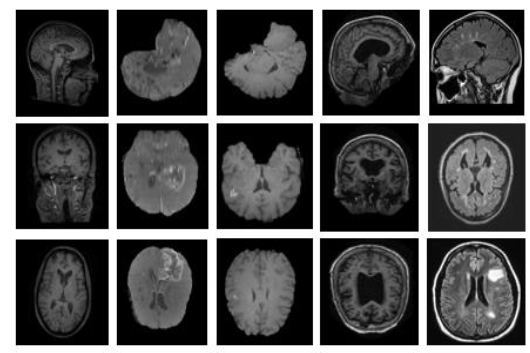}
\centering
\caption{We trained the proposed network on five different categories of brain MRI. The 1st column shows healthy brain in sagittal, coronal and axial section. The 2nd and 3rd columns show high and low grad glioma, while 4th and 5th columns present some brain images on Alzheimer and multiple sclerosis.}
\label{fig_5cat_classification}
\end{figure}

In the experiment we applied real patient data from five popular benchmarks to evaluate the proposed methods.
For classification task we totally compiled 1500 MRI images with label of healthy, tumor-HGG, tumor-LGG, Alzheimer and multiple sclerosis. We consider 20\% of the data for testing and 80\% for training. IXI dataset~\cite{urlixi} contains 600 MRI images from normal, healthy subjects. The MRI image acquisition protocol for each subject includes six modalities, from which we have used T1, T2, PD, MRA images.
The first column of Figure~\ref{fig_5cat_classification} shows the healthy brain images from IXI dataset in the sagittal, coronal and axial sections.
The BraTS2016 benchmark~\cite{urlbrats,Menze2014} prepared the data in two part of High and Low Grade Glioma (HGG/LGG) Tumor. 
All images have been aligned to the same anatomical template and interpolated to 1 mm, 3 voxel resolution.
The training dataset consists of 220 HGG and 108 LGG MRI images which for each patient T1, T1contrast, T2, FLAIR and ground truth labeled by medical experts have been provided.
Alzheimer disease dataset\footnote{\url{http://www.oasis-brains.org/}} comes from Open Access Series of Imaging Studies (OASIS). 
The dataset consists of a cross-sectional collection of 416 subjects aged from 18 to 96. 
For each subject, 3 or 4 individual T1-weighted MRI scans were obtained in single scan sessions. 
18 MRI images with multiple sclerosis from ISBI challenges 2008\cite{urlMS} have also been applied in the classification task. 
ISLES benchmark 2016~\cite{urlisles} (Ischemic Stroke Lesion Segmentation) comes from MICCAI challenge in two part, by which we used only SPES dataset with 30 brain images with 7 modalities in our task. 
An visual overview of the applied datasets can be found in Figure~\ref{fig_5cat_classification}.

\begin{figure}[!t]
\center
\includegraphics[width=0.7\textwidth]{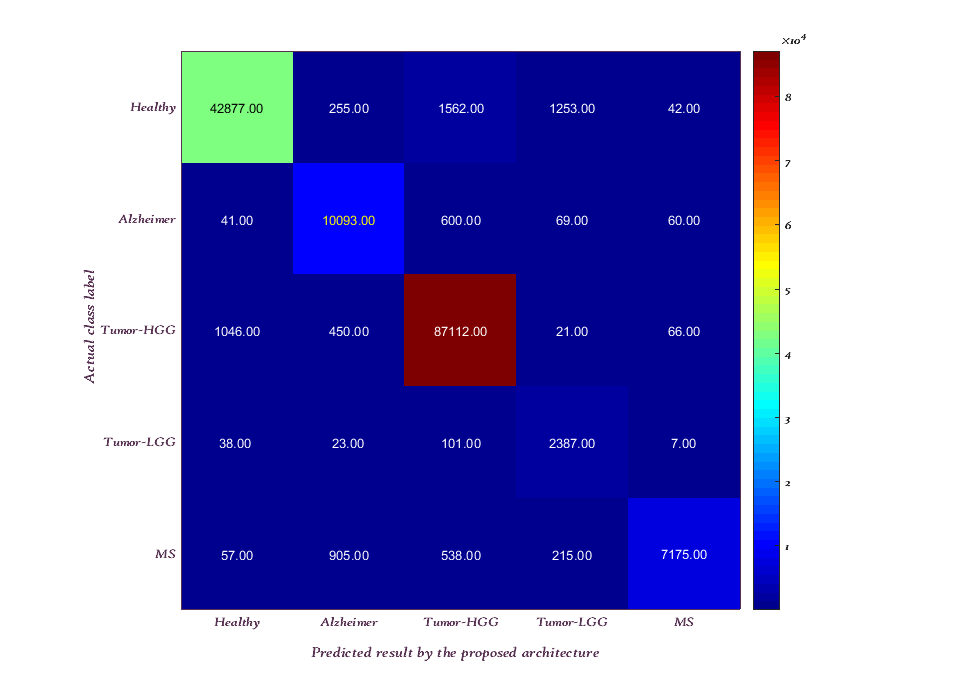}
\caption{The confusion matrix of the classification results. X-axis shows predicted results, where Y-axis gives the actual labels.}
\label{confusion_matrix}
\end{figure}
 
\begin{table}[]
\centering
\caption{Brain lesions classification performance of the re-designed ResNet architecture using l2-norm unit. The involved classes include healthy, tumor-HGG, tumor-LGG, Alzheimer and multiple sclerosis. The last two rows show the comparison results to the most recent methods.}
\label{table-classify-result}
\begin{tabular}{|l|l|l|l|l|l|l|}
\hline
        ~ & Total MRI & Accuracy & Sensitivity & Specificity & Recall & Kappa \\
\hline
        Our method & 1500 & 95.308\% & 0.91 & 0.87& 87.65 & 0.92  \\ 
        Justin S et al.~\cite{doi2254195}  & 191 & 91.43\% & - & - & - & -  \\
        Abbadi et al.~\cite{el2017brain} & 50 & 94\% & 0.85 & 0.87& - & -  \\
\hline
\end{tabular}
\end{table}

\begin{figure}
\center
\includegraphics[width=1\textwidth]{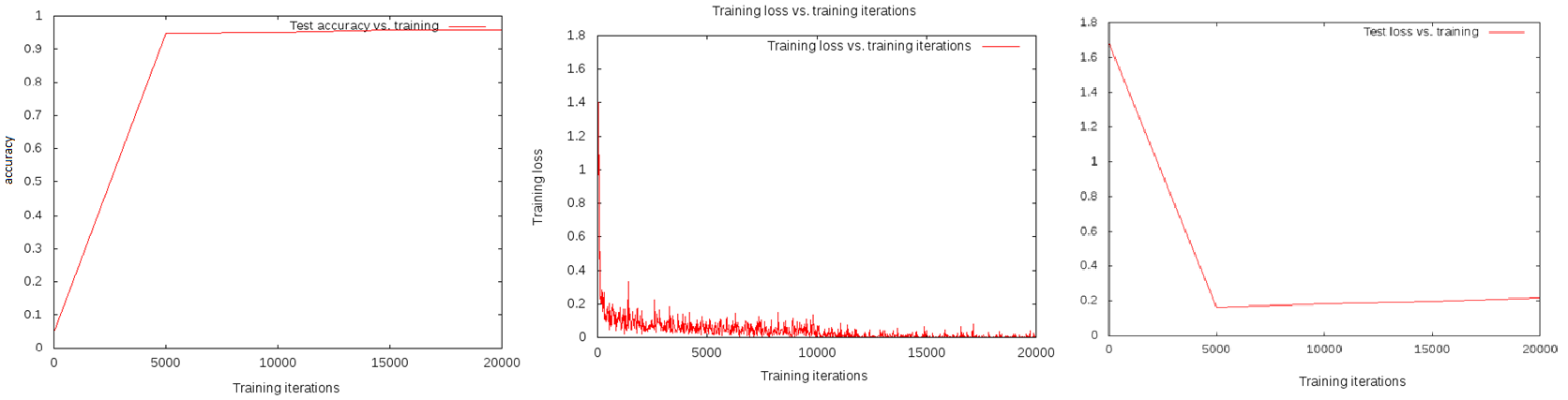}
\caption{Learning curves of brain lesions classification}
\label{fig_kurve}
\end{figure}

Because the MRI volumes in the BraTS and ISLES datasets do not possess an isotropic resolution, we prepared 2D slices in sagittal, axial and coronal view. 
As mentioned by Havaei et al.~\cite{havaei2017brain}, unfortunately brain imaging data are rarely balanced due to the small size of the lesion compared to the rest of the brain. 
For example the volume of a stroke is rarely more than 1\% of the entire brain and a tumor (even large glioblastomas) never occupies more than 4\% of the brain. 
Training a deep network with imbalanced data often leads to very low true positive rate since the system gets to be biased towards the one class that is over represented. 
To overcome this problem we have chosen volume of MRI with lesions, and augmented training data by using horizontal and ventricle flipping, multiple scaling. 
By using a re-designed ResNet architecture described in section~\ref{methodology}, we achieved over 95\% classification accuracy as shown in Table~\ref{table-classify-result}, while Figure~\ref{confusion_matrix} demonstrates the confusion matrix of the classification result. 
We also compared our result with the most recent deep learning based approaches as shown in Table~\ref{table-classify-result}, where the reference method also used IXI, OASIS datasets.
Figure~\ref{fig_kurve} shows learning curves of testing accuracy, training and testing losses during the training process. 

For brain lesions detection experiment we applied both BraTS and ISLES datasets. We used 70\% of the data for training, 10\% for validation and 20\% for testing. 
It is expected that more generalized features could be able to learned from multiple modalities, and the testing accuracy based on more generalized features should be gained.
The brain lesions detection results from Table~\ref{table_detection} proved our assumption, where better detection results were achieved by increasing the data modalities in the model training. 
The detection result can be improved by 20\% in BraTS and 30\% in ISLES dataset.

\begin{table}[!t]
\centering
\caption{Dice Similarity Coefficient (DSC) results (for brain lesions detection performance measurement) on the BraTS2016 and ISLES2016 dataset by using incremental modalities. F/D column means the FLAIR modality in BraTS dataset and DWI modality in ISLES dataset.}
\label{table_detection}
\begin{tabular}{|l|l|l|l|l|l|}
\hline
        T1 & T1c & T2 &F/D & Dice-BraTS16 &  Dice-ISLES  \\
        \hline
         x & -  & - & -  & 61.8 \% & 42\% \\
         - & x & - & - & 33.76 \% & 27\% \\
         - & - & x & - & 36.7 \% & 39.98\% \\
         - & - & - & x & 73.38 \% & 50.71\% \\
         - & x & x & x & 81.53 \% & 54.23\% \\
         x & x & - & x & 82.6 \% & 54.67\% \\
         x & - & x & x & 83.19 \% & 53.09\% \\
         x & x & x & - & 82.73 \% & 54.7\% \\
         x & x & x & x & 83.53 \% & 56.87\% \\        
\hline
\end{tabular}
\end{table}

\begin{table}[]
\centering
\caption{Evaluation result of the detection network with and without l2-norm unit, which demonstrates the performance gains by using l2-norm unit.}
\label{table_compare_l2norm_before_after}
\begin{tabular}{|l|l|l|l|}
\hline
        ~ &Dice(without l2-norm unit) & Dice(with l2-norm unit)  \\\hline
        BraTS16 & 72\% & 83.53\%  \\ 
        ISLES16 & 53.65\% & 56.87 \\  
\hline      
\end{tabular}
\end{table}

\begin{figure}[!t]
\center
\includegraphics[width=0.85\textwidth]{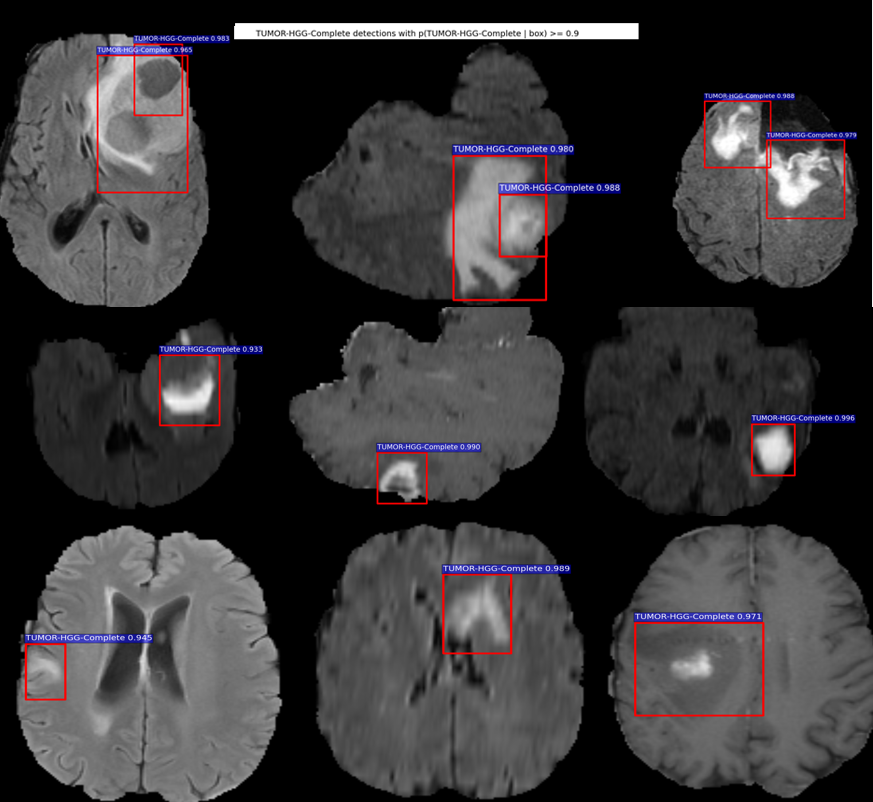}
\caption{Visual results of our brain lesions detection approach on Axial, Coronal and Sagittal views. The subjects are selected from the validation set.}
\label{detection_result}
\end{figure}

From Table~\ref{table_detection}, we can also infer that the FLAIR modality is the most relevant one for identifying the complete tumor(Dice: 73.38\%),
However in ISLES benchmark we don't have this modality and it is justified less accuracy on this category. 
It motivated us to work on generating the missing modalities in the future. 
The subjects in Figure~\ref{detection_result} are from our testing set, for which the model is not trained on, the detection results from these subjects could give a good estimation of the model performance.

Table~\ref{table_compare_l2norm_before_after} demonstrates the evaluation results of the detection architectures with and without l2-norm unit.
From which we can easily realize the superior ability of the proposed l2-norm operator.
We are able to improve the detection performance significantly on both datasets by using this novel operator.

\section{Conclusion} \label{conclusion}

In this paper, we explored two important clinical tasks: brain lesions classification and detection.
We proposed end-to-end trainable approaches based on state-of-the-art deep convolutional neural networks.
We implemented a novel pooling operator: l2-norm unit which can effectively generalize the network, and make the learned model more robust.
The applicability, model accuracy and generalization ability have been evaluated by using a set of publicly available datasets.
As the future work we will further investigate the automatic segmentation of tumor regions based on the detection results.

\bibliographystyle{splncs03}
\bibliography{references}

\end{document}